\documentclass[journal,twoside,web]{ieeecolor}
\usepackage{generic}
\usepackage{cite}
\usepackage{amsmath,amssymb,amsfonts}
\usepackage{algorithm}
\usepackage{algorithmic}
\usepackage{graphicx}
\usepackage{epstopdf}
\usepackage{textcomp}
\usepackage{mathtools}
\makeatletter
\let\NAT@parse\undefined
\makeatother
\usepackage[colorlinks,linkcolor=blue,citecolor=blue,urlcolor=blue]{hyperref}

\def\BibTeX{{\rm B\kern-.05em{\sc i\kern-.025em b}\kern-.08em
T\kern-.1667em\lower.7ex\hbox{E}\kern-.125emX}}

\makeatletter
\def\ps@titlepagestyle{\def\@oddfoot{}\def\@evenfoot{}%
  \def\@oddhead{%
    \vbox{%
      \hbox to \textwidth{\hfill\scriptsize\textsf{\thepage}}%
      \vspace*{2pt}%
      \hbox to \textwidth{\color{subsectioncolor}\leaders\hrule height 1pt\hfill}%
    }%
  }%
  \def\@evenhead{}%
}
\makeatother

\begin{document}

\title{Agentic Neuro--Symbolic Planning and Commissioning for Human-in-the-Loop Industrial Robotics with Digital Twins}

\author{Zhihao Liu,~\IEEEmembership{Member,~IEEE}, Victor Nan Fernandez-Ayala,~\IEEEmembership{Graduate Student Member,~IEEE}, Tianyu Wang, Qiang Qin, Xi Vincent Wang, Dimos V. Dimarogonas,~\IEEEmembership{Fellow,~IEEE}, and Lihui Wang
    \thanks{This work was supported by Swedish Digital Futures' Industrial Innovation project: Towards Safe Smart Construction: Algorithms, Digital Twins and Infrastructures (VF 2020-0315).}
    \thanks{Corresponding author: Zhihao Liu.}
    \thanks{Zhihao Liu, Tianyu Wang, Qiang Qin, Xi Vincent Wang, and Lihui Wang are with the Department of Production Engineering, School of ITM, Royal Institute of Technology (KTH), 100 44 Stockholm, Sweden (Email: {\tt\small zhihaoliu@ieee.org, qiangq@kth.se, tianyuwa@kth.se, wangxi@kth.se, lihuiw@kth.se}). }
  \thanks{Victor Nan Fernandez-Ayala, and Dimos V. Dimarogonas are with the Department of Decision and Control Systems, School of EECS, Royal Institute of Technology (KTH), 100 44 Stockholm, Sweden (Email: {\tt\small vnfa@kth.se, dimos@kth.se}).}
\thanks{During the preparation of this work the author(s) used Anthropic Opus 4.7 in order to help prepare the initial draft of this manuscript, design code for experiments, and polish writing. After using this tool/service, the author(s) reviewed and edited the content as needed and take(s) full responsibility for the content of the published article.}
\thanks{This work has been submitted to the IEEE for possible publication. Copyright may be transferred without notice, after which this version may no longer be accessible.}
}
\IEEEaftertitletext{\vspace{-3.5\baselineskip}}

\maketitle

\begin{abstract}
  Flexible robotic automation requires systems that interpret operator intent, verify physical feasibility, and recover from execution failures across both the planning and execution stages. This paper proposes an agentic neuro-symbolic framework for human-in-the-loop industrial robotics, in which LLMs are used for tasks that require language understanding or contextual reasoning, while all verification, sequencing, and execution remain deterministic. The framework adapts the Planner-Generator-Evaluator (PGE) harness pattern from software engineering into a Specifier-Designer-Inspector (SDI) architecture for industrial robotics, combined with LangGraph-based dynamic routing for failure recovery. A two-tier recovery mechanism addresses structure-level replanning through context-aware orchestration and execution-level geometric failures through deterministic recovery skills. A Unity3D digital twin supports human inspection, modification, and re-verification prior to physical execution. Evaluated on natural-language commands across multiple difficulty levels against ten baselines, the proposed method achieves the highest task success. Ablation results confirm that structured command expansion, symbolic verification, selective LLM routing, and recovery skills are each individually necessary.

\end{abstract}


\section{Introduction}
\label{sec:introduction}

\IEEEPARstart{I}{ndustrial} robotics and automation systems increasingly emphasize human-centricity in the transition toward Industry 5.0 and even Industry 6.0 \cite{rojek2025RoleGenerative,lykov2025Industry60}.
Artificial intelligence and digital twins are two emerging technologies for translating high-level human intent into physically executable robot motions \cite{ni2025LargeLanguage, dong2025NextgenerationLLM}.
Recent advances in large language models (LLMs) have demonstrated promising capabilities for generating task plans from natural-language instructions \cite{fan2025EmbodiedIntelligence, kadri2025LLMdrivenAgent,li2026EnvironmentDrivenLLMGuided, tsushima2025TaskPlanninga, wang2026LLMMTMPLarge, xia2025LeveragingLarge}.
However, deploying LLM-generated plans on physical robot systems remains challenging due to the gap between linguistic reasoning and physical constraints, the lack of closed-loop feedback from execution to planning, and the absence of human oversight mechanisms suitable for safety-critical industrial environments.

Existing approaches can be broadly categorized into three groups.
First, classical task and motion planning (TAMP) methods~\cite{garrett2021integrated} provide formal guarantees but require hand-coded domain descriptions and cannot handle natural language input.
Second, LLM-based planning systems such as SayCan~\cite{brohan2023can} and Code as Policies~\cite{liang2023code} use an LLM with predefined prompts but lack execution monitoring, failure recovery, and human-in-the-loop capabilities.
Third, multi-agent LLM frameworks such as AutoGen~\cite{wu2024autogen} and CrewAI~\cite{rashed2025ai} enable collaboration between multiple LLM agents but operate purely in the language domain without connection to physical robot systems or their digital twins.
Furthermore, graph-based orchestration frameworks such as LangGraph~\cite{mandulapalli2025development} enable dynamic, LLM-driven routing between agents, but production experience has revealed that delegating all control-flow decisions to the LLM leads to unpredictable behavior and difficult debugging~\cite{anthropic_harness2025}. This has motivated the Planner-Generator-Evaluator (PGE) harness in software engineering, where deterministic scaffolding wraps LLM calls to ensure reliability, as demonstrated in tools such as Claude Code~\cite{anthropic_harness2025}.

These observations point to a fundamental tension between neural flexibility and symbolic reliability \cite{hitzler2022NeurosymbolicApproaches, singh2026NeuralsymbolicGrammatical, mao2026BuildingIntelligent}.
LLMs excel at interpreting ambiguous natural language and generating creative spatial arrangements, yet they lack formal guarantees on physical feasibility. An LLM may produce a structure to be assembled by robots that violates gravity constraints, causes collisions, or yields unreachable grasp poses, with no mechanism to detect or correct such errors from within the model itself.
Conversely, classical symbolic methods such as constraint solvers, graph-based planners, and deterministic verifiers can enforce physical laws with mathematical certainty, but they cannot process free-form language or adapt to open-ended task specifications.

This complementarity motivates a neuro-symbolic architecture that combines the two paradigms in a principled manner. In this work, LLMs are used by neural agents for language understanding, structure design generation, and context-aware failure routing, while constraint verification, assembly sequencing, geometric recovery, and execution control are handled by symbolic and reactive agents.

Results show that purely neural baselines without symbolic verification fall below 90\% on open-ended large-scale tasks, while purely symbolic pipelines cannot consume free-form natural-language instructions at all. Only the combined neuro-symbolic configuration sustains full coverage across the benchmark. None of the existing approaches address the complete pipeline from natural language to physical execution with human oversight and closed-loop failure recovery.
The main contributions of this paper are as follows.
\begin{enumerate}
  \item[(i)] A complete human-in-the-loop planning and commissioning pipeline that unifies natural-language task intake, digital-twin supervision, multi-agent reasoning, symbolic verification, and closed-loop robot execution, bridging AI inference and human oversight at every decision point of industrial robot systems.
  \item[(ii)] A Specifier-Designer-Inspector (SDI) multi-agent orchestration that adapts the Planner-Generator-Evaluator harness pattern from software engineering to industrial robots, coordinating heterogeneous neural and symbolic agents through deterministic scaffolding plus LLM-driven dynamic routing, with the Inspector replacing the LLM-based critic with a symbolic verifier for formal feasibility guarantees.
  \item[(iii)] A two-tier closed-loop failure recovery mechanism not addressed in prior LLM-based planning systems: context-aware orchestration drives layout replanning at the design tier, while domain-specific geometric recovery skills handle motion-planning failures at the execution tier, both coupled to the digital twin for automatic re-verification and human-in-the-loop correction.

\end{enumerate}

\section{Formulation}
\label{sec:formulation}

An LLM is modeled as an autoregressive mapping parameterized by $\boldsymbol{\theta}$ that, given a prompt $s \in \Sigma^*$, produces a token sequence $y = f_{\boldsymbol{\theta}}(s)$ through one complete decoding pass. We use this notation $f_{\boldsymbol{\theta}}(\cdot)$ throughout the paper to denote a single LLM inference call.

Without loss of generality, we formalize the human-in-the-loop robot task-and-motion planning problem for industrial robotic automation, where natural-language instructions, interactive digital twins, symbolic task reasoning, and continuous motion planning are tightly integrated.
Given a natural language instruction $l \in \mathcal{L}$ and optional human adjustment input $I_h \in \mathcal{I}_{\text{human}}$ from digital twin, find a feasible robot trajectory $Q \in \mathcal{Q}$ that assembles the target structure while satisfying all physical and kinematic constraints:
\begin{equation}
  \Pi: \mathcal{L} \times \mathcal{I}_{\text{human}} \rightarrow \mathcal{Q}
  \label{eq:problem}
\end{equation}

The problem involves two types of human input:
\begin{itemize}
  \item $\mathcal{L} = \Sigma^*$: Natural language instruction space, where humans specify high-level assembly goals (e.g., ``build a 3-cube tower'')
  \item $\mathcal{I}_{\text{human}}$: Adjustment input space in the digital twin, where humans refine object configurations through direct manipulation after visual inspection
\end{itemize}

The output space is:
\begin{itemize}
  \item $\mathcal{Q} = \prod_{r=1}^{R} \mathbb{R}^{d_r \times T_r}$: Joint trajectory space for $R$ robot manipulators, where $d_r$ is the number of joints for robot $r$ and $T_r$ is its trajectory length
\end{itemize}

We distinguish between planned and executed trajectories:
\begin{itemize}
  \item $Q^{\text{plan}} \in \mathcal{Q}$: Planned trajectory from commissioning via digital twin
  \item $Q^{\text{exec}} \in \mathcal{Q}$: Executed trajectory recorded from remote monitoring via digital twin
\end{itemize}

This problem decomposes into the following sub-problems.

\textbf{Sub-problem 1}: Target Structure Design Generation. Given $l \in \mathcal{L}$, generate a structure configuration $C \in \mathcal{C}$:
\begin{equation}
  C = \{c_1, c_2, \ldots, c_n\}, \quad c_i = (id_i, p_i, q_i, s_i)
  \label{eq:layout_def}
\end{equation}
where $id_i$ is the ID of any object, $p_i \in \mathbb{R}^3$ is position, $q_i \in SO(3)$ is orientation, and $s_i \in \mathbb{R}^3_+$ is the object size.
This step corresponds to mapping linguistic intent into a geometric representation in the digital twin.

\textbf{Sub-problem 2}: Human Adjustment. After commissioning in the digital twin, the human may refine the generated layout.
Given $C \in \mathcal{C}$ and human input $I_h \in \mathcal{I}_{\text{human}}$, a refined configuration is obtained as
\begin{equation}
  C' = C \oplus I_h,
  \label{eq:layout_def_after_human}
\end{equation}
where $\oplus$ denotes the application of human modifications, including translating or rotating existing objects, inserting new objects, and removing objects from the configuration.

\textbf{Sub-problem 3}: Task Sequence Planning. Given the refined configuration $C' = \{c_1, c_2, \ldots, c_n\}$ as defined in (\ref{eq:layout_def_after_human}), find an assembly sequence
\begin{equation}
  \mathcal{S} = (\sigma_1, c_1)(\sigma_2, c_2) \cdots (\sigma_n, c_n)
  \label{eq:sequence}
\end{equation}
where $\sigma_i$ denotes the procedural order index of $c_i$, such that the overall solution is feasible in the sense of Section~\ref{sec:constraints}.

\textbf{Sub-problem 4}: Motion Planning and Commissioning through Digital Twin. Given an assembly sequence $\mathcal{S}$, compute collision-free joint trajectories $Q^{\text{plan}} \in \mathcal{Q}$ for the robot manipulators.
Motion planning is executed sequentially along $\mathcal{S}$ using the current robot state at each step.

\textbf{Sub-problem 5}: Motion Monitoring and Validation through Digital Twin. This sub-problem ensures execution reliability and safety by quantitatively checking whether physical robot motion remains consistent with the planned trajectory and by providing a criterion for triggering feedback actions (e.g., warning, pause, or replanning). Given the executed trajectory $Q^{\text{exec}}$ recorded from robot joint encoders, we verify that execution conforms to the plan:
\begin{equation}
  V_{\text{traj}}: Q^{\text{plan}} \times Q^{\text{exec}} \rightarrow \{\text{pass}, \text{fail}\}, \quad Q^{\text{plan}}, Q^{\text{exec}} \in \mathcal{Q}
  \label{eq:traj_verify}
\end{equation}

The validation is based on the trajectory deviation metric:
\begin{equation}
  \delta(Q^{\text{plan}}, Q^{\text{exec}}) = \max_{r,t} \|q_r^{\text{plan}}(t) - q_r^{\text{exec}}(t)\|_2
  \label{eq:traj_deviation}
\end{equation}
where $q_r(t)$ denotes the joint configuration of robot $r$ at time $t$.

The trajectory is validated as follows:
\begin{equation}
  V_{\text{traj}}(Q^{\text{plan}}, Q^{\text{exec}}) =
  \begin{cases}
    \text{pass} & \text{if } \delta(Q^{\text{plan}}, Q^{\text{exec}}) \leq \epsilon \\
    \text{fail} & \text{otherwise}
  \end{cases}
  \label{eq:traj_check}
\end{equation}
where $\epsilon$ is the acceptable deviation threshold.

\label{sec:constraints}

A structure configuration $C$ and its assembly sequence $\mathcal{S}$ are jointly \textit{feasible} if they satisfy the following six constraints, which together define the constraint set $\mathcal{K}$ referenced throughout the methodology. Two objects $c_i$ and $c_j$ are \textit{in contact} when their axis-aligned bounding boxes meet along all three axes within a tolerance $\epsilon$:
\begin{equation}
  \text{contact}(c_i, c_j) \triangleq \bigwedge_{d \in \{x,y,z\}} \left( |p_i^d - p_j^d| \leq \frac{s_i^d + s_j^d}{2} + \epsilon \right),
  \label{eq:contact}
\end{equation}
where $p_i$ and $s_i$ are defined in (\ref{eq:layout_def}). This predicate is used in Constraints~1 and~4 below.

\textit{Constraint 1 (Gravity):} Objects must be supported from below. For objects in contact $c_i, c_j$:
\begin{equation}
  \text{contact}(c_i, c_j) \land (p_i^z < p_j^z) \Rightarrow \sigma_i < \sigma_j
  \label{eq:gravity}
\end{equation}
where superscript $z$ indicates the Z axis, $\sigma_i$ is the placement order index of $c_i$.

\textit{Constraint 2 (Reachability):} All objects must lie within the robot workspace. For tractability, we use a cuboid workspace as an example, where $x$, $y$, and $z$ indicate the size:
\begin{equation}
  \forall c_i: p_i \in \mathcal{W}, \quad \mathcal{W} = [x_{\min}, x_{\max}] \times [y_{\min}, y_{\max}] \times [z_{\min}, z_{\max}]
  \label{eq:workspace}
\end{equation}

\textit{Constraint 3 (Collision):} Objects cannot overlap:
\begin{equation}
  \forall c_i, c_j: i \neq j \Rightarrow \text{AABB}(c_i) \cap \text{AABB}(c_j) = \emptyset
  \label{eq:collision}
\end{equation}
where AABB denotes the \emph{Axis-Aligned Bounding Box}, i.e., the minimum box in the world frame that encloses an object.

\textit{Constraint 4 (Contact):} Non-ground objects must have structural contact:
\begin{equation}
  \forall c_i: p_i^z > 0 \Rightarrow \exists c_j: \text{contact}(c_i, c_j)
  \label{eq:contact_constraint}
\end{equation}
This constraint enforces static support (no suspension), whereas Constraint~1 enforces the corresponding placement order under gravity.

\textit{Constraint 5 (Gripping):} At least one feasible gripping exists:
\begin{equation}
  \forall c_i: \exists \beta \in \{x, y\}: \neg\text{blocked}(c_i, \beta, \mathcal{S}_{<\sigma_i})
  \label{eq:gripping}
\end{equation}
where $\mathcal{S}_{<\sigma_i}$ denotes the already placed objects.
This condition is critical for practical execution with parallel-jaw (two-finger) grippers, which require at least one collision-free approach direction to grasp an object.

\textit{Constraint 6 (Sequence):} Assembly order respects physical dependencies:
\begin{equation}
  \forall c_i, c_j: \text{supports}(c_i, c_j) \Rightarrow \sigma_i < \sigma_j
  \label{eq:sequence_constraint}
\end{equation}
This constraint primarily captures product-level assembly precedence defined at design time, i.e., which component should be assembled first and which should follow. In contrast, Constraints~1 and~4 focus on physical feasibility (gravity and static support).

\section{Methodology}
\label{sec:methodology}

\subsection{System Workflow}
\label{sec:system_pipeline}

\begin{figure}[t]
  \centering
  \includegraphics[width=\columnwidth]{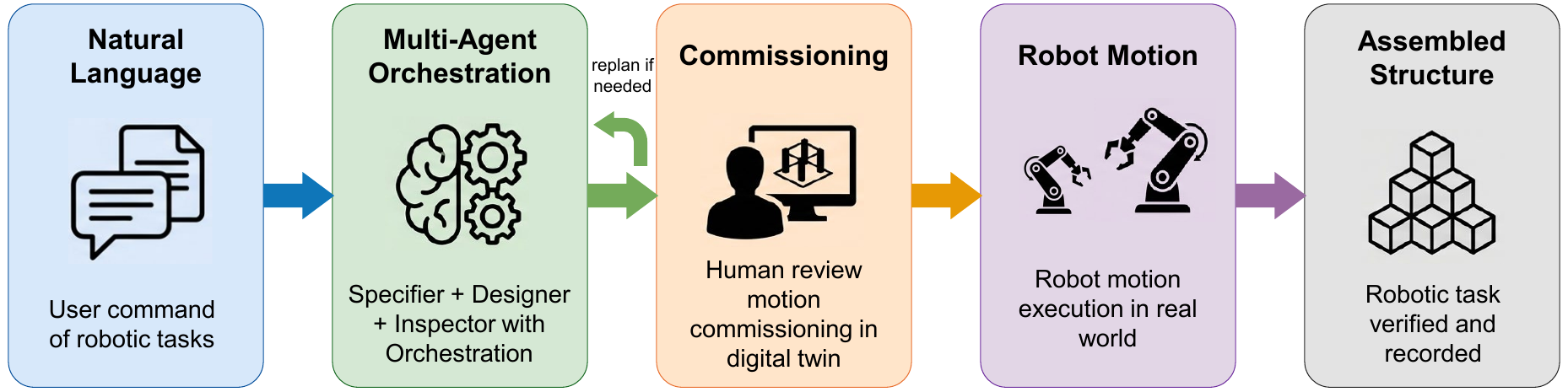}
  \caption{System workflow.}
  \label{fig:pipeline}
\end{figure}

The proposed human-in-the-loop task-and-motion planning and commissioning pipeline is summarized by the following operator chain and Fig.~\ref{fig:pipeline}:
\begin{equation}
  \pi:~ \mathcal{L} \xrightarrow{f_{\boldsymbol{\theta}}(\cdot;\,\phi(E))} \mathcal{C}
  \xrightarrow{h_{\text{DT}}} \mathcal{C}' \xrightarrow{g_{\text{sym}}} \mathcal{C}^*
  \xrightarrow{\tau_{\text{task}}} \mathcal{T}
  \xrightarrow{\rho_{\text{motion}}} Q^{\text{plan}}.
  \label{eq:pipeline}
\end{equation}
To close the loop during deployment, the execution and monitoring processes are defined as
\begin{equation}
  \begin{aligned}
    \pi_{\text{exec}}:~
    & Q^{\text{plan}} \xrightarrow{\text{execute}} Q^{\text{exec}}
    \xrightarrow{V_{\text{traj}}} \{\text{pass},\text{fail}\}, \\
    & Q^{\text{plan}} \xrightarrow{\eta_{\text{DT}}} \mathcal{Z}, \\
    & Q^{\text{exec}} \xrightarrow{\eta_{\text{real}}} \mathcal{Z}.
  \end{aligned}
  \label{eq:pipeline_feedback}
\end{equation}
where $\pi_{\text{exec}}$ denotes the deployment-time execution and monitoring operator, $\eta_{\text{DT}}$ indicates the commissioning of planned trajectory in digital twin, $\eta_{\text{real}}$ monitors the executed trajectory in digital twin, and $\mathcal{Z}$ denotes the digital space in digital twin.

Together, (\ref{eq:pipeline}) and (\ref{eq:pipeline_feedback}) define a closed-loop workflow from natural-language intent to physical execution under digital-twin commissioning and monitoring. The pipeline is organized into four functional modules, summarized below.

Module 1: Structure design generation (\ref{sec:agentic_design}). The natural-language command is expanded into a structured specification, used to generate a candidate structure, and verified against physical constraints, with violations fed back for targeted regeneration until feasibility is reached.

Module 2: Human-in-the-loop digital twin (\ref{sec:human_centered}). The verified layout is rendered in the digital twin for the operator to accept, modify, or reject. Any edits are automatically re-verified before proceeding.

Module 3: Symbolic sequencing and task planning (\ref{sec:symbolic}). A graph-based symbolic planner derives an assembly order respecting gravity, support, and gripping feasibility, which is then compiled into a robot-level action sequence.

Module 4: Motion planning, previewing, and execution (\ref{sec:task_motion}). Collision-free trajectories are computed, previewed in the digital twin, and executed on the physical system. Measured trajectories are compared against planned ones for in-loop monitoring and recovery.

Although presented in order for clarity, the four modules share three cross-cutting capabilities. Symbolic constraint reasoning appears in both Module~1 (feasibility verification of LLM-generated structures) and Module~3 (gravity, support, and gripping constraints during sequencing). The digital twin serves as a unified commissioning interface across Modules~2 and~4, rendering structures, task sequences, and planned/executed trajectories for human review and in-loop monitoring. Multi-agent orchestration (\ref{sec:multi_agent}) spans the entire pipeline: heterogeneous agents coordinate sequencing, task compilation, execution, and recovery, and failures in Modules~3--4 loop back to Module~1 via LLM-driven routing.

\subsection{Agentic System and Orchestration}
\label{sec:agents}
\label{sec:agent_types}
\label{sec:multi_agent}

\begin{figure}[t]
  \centering
  \includegraphics[width=\columnwidth]{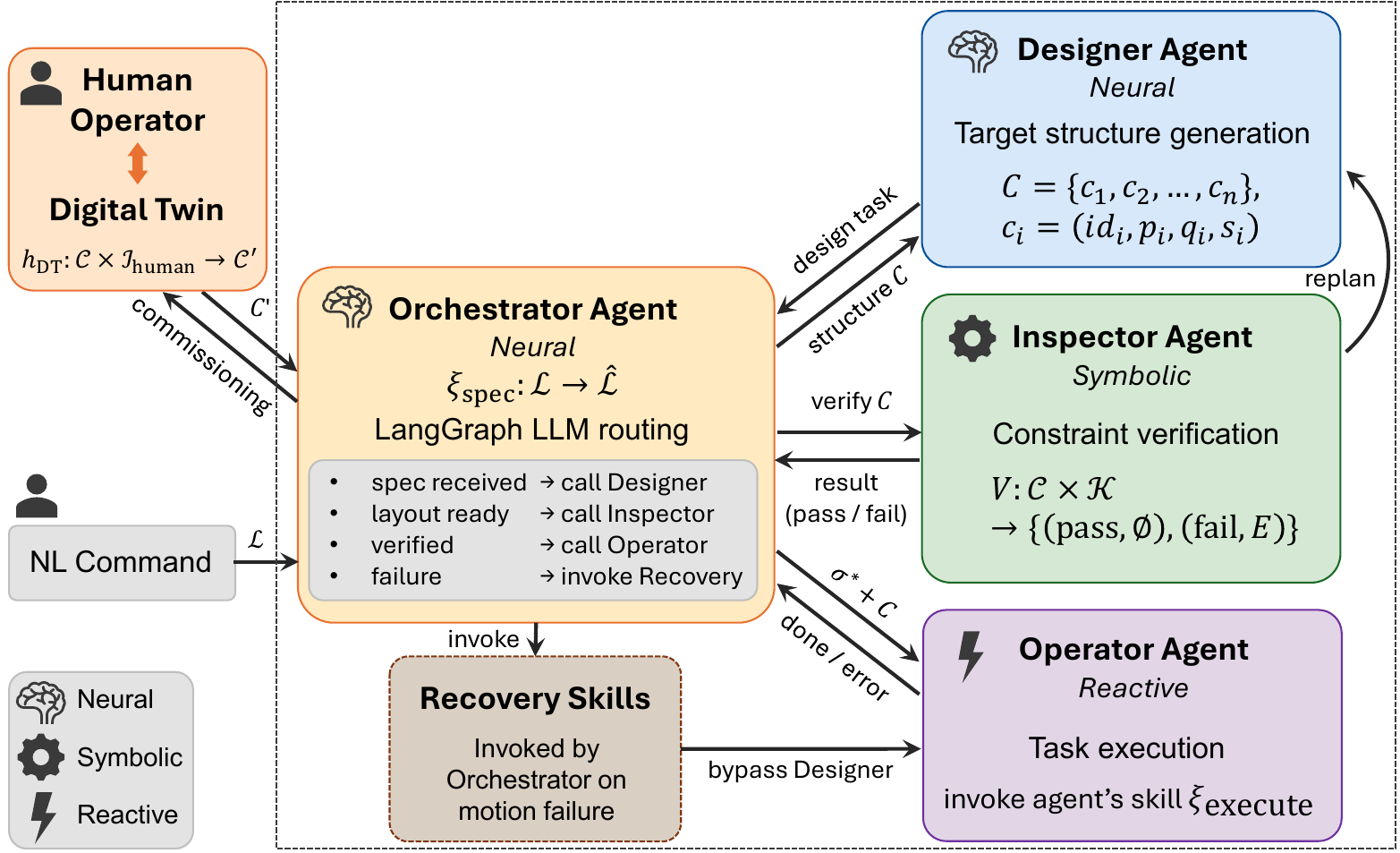}
  \caption{Data flow of the multi-agent orchestration.}
  \label{fig:orchestration}
\end{figure}

We adopt a heterogeneous agentic architecture to cover the full pipeline from structure design to physical execution with human oversight.
The system consists of four computational agents and one human operator, as shown in Fig.~\ref{fig:orchestration}.

An agent $a_i$ is defined as a tuple $a_i = (\text{type}_i, \Xi_i, \pi_i)$, where $\text{type}_i \in \{\text{Neural}, \text{Symbolic}, \text{Reactive}\}$ denotes the computational paradigm, $\Xi_i$ is the agent's skill set, and $\pi_i$ is the decision policy.
The human operator $\mathcal{H}$ interacts with the system through the digital twin and is not modeled as an agent but as an external decision-maker.

The three agent types follow the design principle that LLMs are used only where necessary.
\textit{Neural} agents handle language understanding and contextual reasoning.
\textit{Symbolic} agents use deterministic algorithms for verification and sequencing.
\textit{Reactive} agents interface directly with the robot system for real-time execution.

All agents share a unified skill abstraction.
\label{sec:skill_abstraction}
Formally, a skill is a self-contained tool-callable module that encapsulates a specific verification or planning capability. It is represented as a tuple $\xi=(\textit{name}, \textit{desc}, \Theta, f_\xi)$, in which $\textit{name}\in\Sigma^*$ denotes a unique identifier, $\textit{desc}\in\Sigma^*$ denotes a natural-language description exposed to the LLM for tool selection, $\Theta$ denotes a JSON Schema that specifies the input parameter space, and $f_\xi:\Theta\rightarrow\mathcal{R}$ denotes the execution function that maps validated inputs to a structured result.
The result space $\mathcal{R}$ is defined as
\begin{equation}
  \mathcal{R} = \{\textsc{Success}, \textsc{Failure}, \textsc{Partial}\} \times \mathcal{D} \times 2^{\Sigma^*}
  \label{eq:skill_result}
\end{equation}
where the first component is the execution status, $\mathcal{D}$ is the skill-specific result data (e.g., a generated layout, an assembly sequence, or a trajectory record), and the third component is a set of error descriptions returned on failure or partial completion.
Each result $r \in \mathcal{R}$ is serialized in JSON and returned to the LLM via the \texttt{tool} role message.

\label{sec:registry}
\label{sec:concrete_skills}
A skill registry $\Xi = \{\xi_1, \ldots, \xi_m\}$ \label{eq:registry} manages all available skills.
It generates tool schemas $\textsc{Tools}(\Xi)$ \label{eq:tools} and dispatches tool calls as $\textsc{Dispatch}(\Xi, n, a) = f_{\xi}(a)$ \label{eq:dispatch}.
Across the system, skills are organized into four categories:
(i)~\textit{composite verification} ($\xi_{\text{verify}}$), which wraps the full constraint verifier $V(C, \mathcal{K})$;
(ii)~\textit{granular verification} ($\xi_{\text{ws}}, \xi_{\text{grav}}, \xi_{\text{grip}}, \xi_{\text{col}}$), which isolate individual constraint checks for targeted diagnosis;
(iii)~\textit{planning skills} ($\xi_{\text{seq}}, \xi_{\text{task}}$), which invoke assembly sequencing and task generation; and
(iv)~\textit{orchestrating skills} ($\xi_{\text{spec}}, \xi_{\text{flatten}}, \xi_{\text{shift}}$), comprising the Specifier for command expansion and domain-specific recovery skills for geometric failure correction. 

\textbf{Orchestrator Agent (Neural)} decomposes user commands into execution phases, schedules agents, and handles failures through replanning or escalation to the human operator.
It implements the proposed SDI harness, inspired by the PGE pattern~\cite{anthropic_harness2025}, combined with LangGraph-based dynamic routing~\cite{mandulapalli2025development}.
The SDI Specifier ($\xi_{\text{spec}}$) and Inspector verification provide deterministic scaffolding where reliability is critical, whereas LangGraph routing is reserved for failure recovery where context-awareness improves outcomes.
Domain-specific recovery skills ($\xi_{\text{flatten}}, \xi_{\text{shift}}$) address geometric failures that neither fixed rules nor LLM replanning can reliably resolve.
Concretely, it implements a LangGraph-based state machine over the phase space $\mathcal{P}$, consisting of \textsc{Plan}, \textsc{Design}, \textsc{Inspect}, \textsc{HumanReview}, \textsc{PreExecute}, \textsc{Execute}, and \textsc{Replan}, where routing decisions at branch points are made by LLM calls.
In the \textsc{Plan} phase, the Orchestrator invokes $\xi_{\text{spec}}: \mathcal{L} \rightarrow \hat{\mathcal{L}}$ to expand the raw command into a structured specification $\hat{l}$.
In the \textsc{Replan} phase, it first applies recovery skills and falls back to LLM-driven replanning only when needed.
Outputs from recovery skills are routed directly to the Inspector for re-verification. All retry, replanning, and escalation decisions remain with the Orchestrator.

\textbf{Designer Agent (Neural)} generates and modifies layouts from structured specifications, exposing $\xi_{\text{generate}}$ and $\xi_{\text{modify}}$ as skills:
\begin{equation}
  \xi_{\text{generate}}: \hat{\mathcal{L}} \rightarrow \mathcal{C}, \quad
  \xi_{\text{modify}}: \mathcal{C} \times \mathcal{E} \rightarrow \mathcal{C}
  \label{eq:designer_skills}
\end{equation}
where $\hat{\mathcal{L}}$ is the structured specification space produced by the Specifier, and $\mathcal{E}$ denotes error feedback from the Inspector or Operator. The Designer operates within a \emph{generate--verify--repair} loop in which violations are injected back as targeted feedback for the next attempt, and a final safety gate rejects any output that does not satisfy $\mathcal{K}$. Section~\ref{sec:agentic_design} details the implementation of the Designer.

\textbf{Inspector Agent (Symbolic)} performs deterministic constraint verification and assembly sequencing without using an LLM:
\begin{equation}
  \xi_{\text{verify}}: \mathcal{C} \rightarrow \mathcal{R}, \quad
  \xi_{\text{seq}}: \mathcal{C} \rightarrow (\mathcal{C}^*, \sigma^*)
  \label{eq:inspector_skills}
\end{equation}
$\xi_{\text{verify}}$ implements the verifier $V(C, \mathcal{K})$ over the six feasibility constraints defined in Section~\ref{sec:constraints}, returning either pass or a violation set $E \subseteq \mathcal{K}$ that is fed back to the Designer. $\xi_{\text{seq}}$ derives the assembly order $\sigma^*$ via topological sort of a support-induced graph with gripping-feasibility filtering. The detailed implementation is shown in Section~\ref{sec:symbolic}.

\textbf{Operator Agent (Reactive)} serves as the robot execution interface between the symbolic task plan and the physical robot system, both in commissioning in digital twin and real-world execution.
It executes step $k$ through $\xi_{\text{execute}}$, records $Q^{\text{plan}}_k$ and $Q^{\text{exec}}_k$, and reports a structured failure event $e_k = (\text{step}_k, \text{cube\_id}, \text{error\_type}, \text{target\_pose})$ when execution fails.

Agents communicate via a message bus with typed messages $m = (\text{sender}, \text{receiver}, \text{type}, \text{payload}, \text{id})$.
Human interaction is handled separately through direct Orchestrator calls with the digital twin.
When the Operator reports a failure $e_k$, the Orchestrator first tries recovery skills and sends the corrected layout to the Inspector for re-verification.
If recovery fails, it selects among retry, Designer replanning via $\xi_{\text{modify}}$, and human escalation.
The complete pipeline is summarized in Algorithm~\ref{alg:multi_agent}.

\begin{algorithm}[t]
  \caption{Multi-Agent Orchestration with SDI Harness}
  \label{alg:multi_agent}
  \begin{algorithmic}[1]
    \REQUIRE $l \in \mathcal{L}$, agents $\{a_{\text{orch}}, a_{\text{des}}, a_{\text{insp}}, a_{\text{op}}\}$, human interface $h_{\text{DT}}$
    \ENSURE Motion task completed or \textsc{Failure}
    \STATE $\hat{l} \gets a_{\text{orch}}.\xi_{\text{spec}}(l)$ \COMMENT{Specifier expands command}
    \STATE $C \gets a_{\text{des}}.\xi_{\text{generate}}(\hat{l})$ \COMMENT{Designer generates layout}
    \STATE $(C^*, \sigma^*) \gets a_{\text{insp}}.\xi_{\text{verify\_and\_seq}}(C)$ \COMMENT{Inspector verifies}
    \WHILE{verification failed}
    \STATE $C \gets a_{\text{des}}.\xi_{\text{modify}}(C, \text{errors})$ \COMMENT{Designer replans}
    \STATE $(C^*, \sigma^*) \gets a_{\text{insp}}.\xi_{\text{verify\_and\_seq}}(C)$ \COMMENT{Re-verify}
    \ENDWHILE
    \STATE $C' \gets h_{\text{DT}}.\text{request\_review}(C^*)$ \COMMENT{Human reviews in digital twin}
    \IF{human modified layout}
    \STATE $(C^*, \sigma^*) \gets a_{\text{insp}}.\xi_{\text{verify\_and\_seq}}(C')$ \COMMENT{Re-verify}
    \ENDIF
    \STATE $h_{\text{DT}}.\text{pre\_execute\_preview}(\sigma^*)$ \COMMENT{Pre-execution digital-twin simulation}
    \FOR{each step $k$ in $\sigma^*$}
    \STATE $(Q^{\text{plan}}_k, Q^{\text{exec}}_k, s_k) \gets a_{\text{op}}.\xi_{\text{execute}}(k)$
    \IF{$s_k = \text{failure}$}
    \STATE $C \gets a_{\text{orch}}.\xi_{\text{recover}}(C, e_k)$ \COMMENT{Apply recovery skill}
    \STATE $(C^*, \sigma^*) \gets a_{\text{insp}}.\xi_{\text{verify\_and\_seq}}(C)$ \COMMENT{Re-verify}
    \IF{re-verification failed}
    \STATE Orchestrator chooses retry, Designer replan, or human escalation
    \ENDIF
    \ENDIF
    \ENDFOR
    \RETURN Motion task completed with trajectory records
  \end{algorithmic}
\end{algorithm}

\subsection{Agentic Target Structure Design Generation}
\label{sec:agentic_design}

This subsection addresses Sub-problem~1 (Target Structure Design Generation): given a structured specification $\hat{l} \in \hat{\mathcal{L}}$, generate a physically feasible configuration $C \in \mathcal{C}$ that satisfies the full constraint set $\mathcal{K}$.

The design module follows a common \emph{generate--verify--repair} paradigm that drives constraint satisfaction. The LLM proposes a candidate layout $C \in \mathcal{C}$; the symbolic verifier $V(C, \mathcal{K})$ either accepts it or returns a violation set $E \subseteq \mathcal{K}$; and the violations are injected back as targeted feedback that constrains the next attempt. This generate--verify--repair cycle repeats until either $V(C, \mathcal{K}) = \text{pass}$ or a step budget $T_{\max}$ is exhausted, and a final safety gate rejects any output that does not satisfy $\mathcal{K}$, so the module returns either a fully feasible $C$ or a \textsc{failure} signal.
Within this paradigm we instantiate two strategies that differ only in how the repair loop is scheduled: an \emph{adaptive constraint-guided} strategy with a fixed control flow and a single LLM call per retry, and an \emph{agentic skill-based} strategy in which the Designer Agent dynamically decides which skills (composite or granular verifiers, layout modifications) to invoke within an episode. We present the two strategies in turn.

\label{sec:lazy_constraints}

Rather than conditioning the LLM on the full constraint set $\mathcal{K}$, we provide only the violated constraints as feedback.
At retry $t$, the prompt is augmented with $\phi(E^{(t-1)})$, yielding $C^{(t)} = f_{\boldsymbol{\theta}}(\hat{l};\,\phi(E^{(t-1)}))$,
where the constraint selector $\phi: 2^{\mathcal{K}} \rightarrow 2^{\mathcal{K}}$ ($|\phi(E)| \ll |\mathcal{K}|$) returns only the constraints relevant to the current errors $E$.
The verifier $V: \mathcal{C} \times \mathcal{K} \rightarrow \{(\text{pass}, \emptyset), (\text{fail}, E)\}$ \label{eq:verify} acts as a deterministic critic.
We build a minimal prompt $s_{\text{minimal}} = [\text{system}] + [\text{few-shot}] + [\text{command}]$ \label{eq:prompt_min} for the first attempt, and $s_{\text{retry}}^{(t)} = s_{\text{minimal}} + [\text{error}] + \phi(E)$ \label{eq:prompt_retry} for retries.
The loop terminates when verification passes or after $T_{\max}$ iterations (Algorithm~\ref{alg:lazy}).

\begin{algorithm}
  \caption{Adaptive Constraint-Guided Inference Loop}
  \label{alg:lazy}
  \begin{algorithmic}[1]
    \REQUIRE $\hat{l} \in \hat{\mathcal{L}}$, $\mathcal{K}$, $T_{\max}$
    \ENSURE feasible $C \in \mathcal{C}$ or \textsc{Failure}
    \STATE $\text{prompt} \gets \text{BuildMinimalPrompt}(\hat{l})$
    \FOR{$t = 1$ \TO $T_{\max}$}
    \STATE $C \gets f_{\boldsymbol{\theta}}(\text{prompt})$
    \STATE $(\text{status}, E) \gets V(C, \mathcal{K})$
    \IF{$\text{status} = \text{pass}$}
    \RETURN $C$
    \ENDIF
    \STATE $\text{prompt} \gets \text{BuildRetryPrompt}(\hat{l}, E, \phi(E), C)$
    \ENDFOR
    \RETURN \textsc{Failure}
  \end{algorithmic}
\end{algorithm}

\label{sec:agentic}

The adaptive constraint-guided loop in Algorithm~\ref{alg:lazy} uses a fixed control flow with one LLM call per retry.
To enable more flexible behavior, the Designer Agent replaces this fixed schedule with an LLM-driven loop that selects skills defined in Section~\ref{sec:agents} through the function-calling interface~\cite{wang2026FunctionCalling}.

\label{sec:agent_loop}
To this end, we maintain a conversation state $\mathcal{M} = (m_1, m_2, \ldots)$ with roles $\{\texttt{system}, \texttt{user}, \texttt{assistant}, \texttt{tool}\}$.
The system message contains the base planning prompt $s_{\text{minimal}}$ together with skill instructions and tool definitions.

At each step $t$, the LLM receives the full conversation context and the tool definitions:
\begin{equation}
  \hat{m}_t = f_{\boldsymbol{\theta}}\!\left(\mathcal{M}_{<t},\; \textsc{Tools}(\Xi)\right)
  \label{eq:agent_call}
\end{equation}

The response $\hat{m}_t$ takes one of two forms:
\begin{enumerate}
  \item A \textbf{tool call} $\hat{m}_t = (\textit{name}, \textit{args})$, in which case the framework executes $r = \textsc{Dispatch}(\Xi, \textit{name}, \textit{args})$, appends the result to $\mathcal{M}$, and continues.
  \item A \textbf{text response}, in which case the framework extracts the final layout $C$ and terminates the loop.
\end{enumerate}

Regardless of the intermediate tool sequence, the framework always performs a final safety check:
\begin{equation}
  C_{\text{final}} \leftarrow
  \begin{cases}
    C & \text{if } V(C, \mathcal{K}) = \text{pass} \\
    \text{FAILURE} & \text{otherwise}
  \end{cases}
  \label{eq:safety_net}
\end{equation}
This guarantees physical feasibility even if the agent skips verification or misuses skills.
The complete agent loop is formalized in Algorithm~\ref{alg:agentic}.
\begin{algorithm}[t]
  \caption{Skill-based Pipeline of Designer Agent}
  \label{alg:agentic}
  \begin{algorithmic}[1]
    \REQUIRE $\hat{l} \in \hat{\mathcal{L}}$, skill registry $\Xi$, step budget $T_{\max}$
    \ENSURE feasible $C \in \mathcal{C}$ or \textsc{Failure}
    \STATE $\mathcal{M} \gets [\texttt{system}: s_{\text{minimal}} + \textsc{SkillInstructions}]$
    \STATE $\mathcal{M} \gets \mathcal{M} \oplus [\texttt{user}: \hat{l}]$
    \FOR{$t = 1$ \TO $T_{\max}$}
    \STATE $\hat{m}_t \gets f_{\boldsymbol{\theta}}(\mathcal{M},\; \textsc{Tools}(\Xi))$ \COMMENT{LLM call with tools}
    \IF{$\hat{m}_t$ contains \texttt{tool\_calls}}
    \STATE $\mathcal{M} \gets \mathcal{M} \oplus [\texttt{assistant}: \hat{m}_t]$
    \FOR{each tool call $(n_j, a_j)$ in $\hat{m}_t$}
    \STATE $r_j \gets \textsc{Dispatch}(\Xi, n_j, a_j)$ \COMMENT{Execute skill}
    \STATE $\mathcal{M} \gets \mathcal{M} \oplus [\texttt{tool}: \textsc{Serialize}(r_j)]$
    \ENDFOR
    \ELSE
    \STATE $C \gets \textsc{ExtractLayout}(\hat{m}_t)$ \COMMENT{Agent done}
    \IF{$C \neq \textsc{null}$}
    \STATE \textbf{break}
    \ENDIF
    \STATE $\mathcal{M} \gets \mathcal{M} \oplus [\texttt{user}: \text{``Output final JSON''}]$
    \ENDIF
    \ENDFOR
    \STATE \COMMENT{Safety check: always verify final output}
    \IF{$C \neq \textsc{null}$ \AND $V(C, \mathcal{K}) = \text{pass}$}
    \RETURN $C$
    \ENDIF
    \RETURN \textsc{Failure}
  \end{algorithmic}
\end{algorithm}

If the safety gate in Algorithm~\ref{alg:agentic} passes, the module returns a layout $C \in \mathcal{C}$ that is guaranteed to satisfy every constraint in $\mathcal{K}$, which is then passed to the remaining pipeline modules; otherwise the episode terminates with a \textsc{failure} signal that is handed back to the Orchestrator for recovery (Section~\ref{sec:agents}).

\subsection{Human-in-the-Loop Verification and Adjustment via Digital Twins}
\label{sec:human_centered}

The digital twin provides a geometry-consistent interface between planning and real-world deployment \cite{liu2026ConstrucTwinDigital, liu2025establishment}.
In our method, it serves as (i) a human-AI interface, (ii) a commissioning environment for designed structure and robotic trajectories, and (iii) a monitoring layer for the physical system.

The human adjustment step is modeled as
\begin{equation}
  h_{\text{DT}}: \mathcal{C} \times \mathcal{I}_{\text{human}} \rightarrow \mathcal{C}',
  \label{eq:human}
\end{equation}
where $\mathcal{I}_{\text{human}}$ denotes human input through visual inspection and direct manipulation in the digital twin.
Typical operations include translating or rotating objects, inserting or removing components, and resolving ambiguous spatial relations.
This step reduces uncertainty before downstream task and motion planning.

Beyond human adjustment, the digital twin also supports commissioning-oriented validation.
Given $C'$ and a planned robotic trajectory $Q^{\text{plan}}$, it can perform collision, reachability, and rule-based safety checks.
These checks provide feedback to either the previous design loop or the later motion planner.

During deployment, execution data can be streamed back to the digital twin, yielding $Q^{\text{exec}}$.
The same environment is then used to quantify deviations and trigger alarms or replanning when thresholds are exceeded (cf. (\ref{eq:traj_deviation}) and (\ref{eq:traj_check})).
Formally, the digital twin provides a shared visualization and monitoring space $\mathcal{Z}$ into which both planned and executed trajectories are projected:
\begin{equation}
  \eta_{\text{DT}}: \mathcal{Q} \rightarrow \mathcal{Z}, \qquad
  \eta_{\text{real}}: \mathcal{Q} \rightarrow \mathcal{Z},
  \label{eq:dt_channels}
\end{equation}
where $\eta_{\text{DT}}$ renders planned trajectories $Q^{\text{plan}}$ for pre-execution commissioning and $\eta_{\text{real}}$ projects executed trajectories $Q^{\text{exec}}$ for in-loop monitoring, closing the feedback loop described in (\ref{eq:pipeline_feedback}).

In the multi-agent pipeline, the human interacts through the digital twin at three points:
(1)~\textit{generated design structure review} after the Inspector verifies a layout, where the human may adjust object positions by direct manipulation (triggering re-verification);
(2)~\textit{pre-execution commissioning and confirmation} where the planned robotic trajectory is visualized in the digital twin before real execution; and
(3)~\textit{failure report} when the Operator reports an execution failure that automatic replanning cannot resolve.
Overall, the digital twin makes the pipeline more interpretable and auditable while reducing trial-and-error on physical industrial robots.

\subsection{Symbolic Assembly Sequencing}
\label{sec:symbolic}

While the agentic design loop may invoke the sequencing skill $\xi_{\text{seq}}$ during target structure design generation, this module applies the same algorithm deterministically to the human-adjusted layout $C'$ to obtain the final assembly order $\sigma^*$.
It enforces the physical dependencies induced by contact and support relations and rejects orders that violate the constraints in Section~\ref{sec:constraints}.

Writing $c_i \prec c_j$ as shorthand for the antecedent of Constraint~1 (\ref{eq:gravity}), we build a directed acyclic graph (DAG) with one node per object and one directed edge per support relation:
\begin{equation}
  G_d = (V, E_d), \quad E_d = \{(v_i \rightarrow v_j) : c_i \prec c_j\}.
  \label{eq:cag}
\end{equation}
We therefore define the symbolic sequencer as
\begin{equation}
  g_{\text{sym}}: \mathcal{C}' \rightarrow \mathcal{C}^*, \quad \sigma^* = \text{TopoSort}(G_d),
  \label{eq:gsym}
\end{equation}
where any topological ordering of $G_d$ respects gravity and structural dependencies.
In practice, we implement this by iterative sink-node removal: each extracted sink yields a disassembly order, which is then reversed to obtain $\sigma^*$.
We use $\mathcal{C}^* \triangleq (C',\sigma^*)$ to denote the sequenced configuration.

Among the valid sink-node orderings, we select one that also satisfies gripping feasibility (\ref{eq:gripping}): a sink node is added only if at least one lateral grip direction $\beta \in \{x,y\}$ remains unblocked.
If no grippable sink node exists, the planner reports failure.
The full procedure is given in Algorithm~\ref{alg:symbolic}.

\begin{algorithm}
  \caption{Symbolic Assembly Sequencing}
  \label{alg:symbolic}
  \begin{algorithmic}[1]
    \REQUIRE $C'$ (layout), $\mathcal{K}$ (constraints)
    \ENSURE $\sigma^*$ (feasible assembly sequence) or \textsc{Failure}
    \STATE $G_d \gets \text{BuildDAG}(C')$ \COMMENT{Edges from $\prec$}
    \STATE $\sigma_{\text{dis}} \gets []$
    \WHILE{$G_d \neq \emptyset$}
    \STATE $\mathcal{F} \gets \{v \in G_d : \text{out-degree}(v) = 0\}$ \COMMENT{Sink nodes}
    \IF{$\mathcal{F} = \emptyset$}
    \RETURN \textsc{Failure} \COMMENT{Circular dependency}
    \ENDIF
    \STATE $c \gets$ first grippable node in $\mathcal{F}$ \COMMENT{cf.~(\ref{eq:gripping})}
    \IF{no grippable node in $\mathcal{F}$}
    \RETURN \textsc{Failure} \COMMENT{No feasible grip}
    \ENDIF
    \STATE append $c$ to $\sigma_{\text{dis}}$; remove $c$ from $G_d$
    \ENDWHILE
    \RETURN $\sigma^* \gets \text{reverse}(\sigma_{\text{dis}})$
  \end{algorithmic}
\end{algorithm}

\subsection{Motion Planning, Commissioning, and Monitoring via Digital Twins}
\label{sec:task_motion}

Given the sequenced configuration $\mathcal{C}^*$, the task generation operator
\begin{equation}
  \tau_{\text{task}}: \mathcal{C}^* \rightarrow \mathcal{T}
  \label{eq:task}
\end{equation}
produces a sequence $\mathcal{T} = (a_1,\ldots,a_K)$ of atomic manipulation actions, each specifying the object, the assigned robot, and the required end-effector poses.
For each action, the robot is assigned according to workspace reachability, and approach and retreat waypoints are generated to avoid collisions.

The motion planning operator then maps task actions to joint-space trajectories:
\begin{equation}
  \rho_{\text{motion}}: \mathcal{T} \rightarrow \mathcal{Q}, \quad
  Q^{\text{plan}} = \bigoplus_{k=1}^{K} f_{\text{mp}}(a_k,\, x_k^{\text{robot}},\, \mathcal{E}_k),
  \label{eq:motion}
\end{equation}
where $x_k^{\text{robot}}$ is the current robot state, $\mathcal{E}_k$ is the collision scene, and $\oplus$ denotes time-ordered trajectory concatenation.
The framework is planner-agnostic. In our implementation, we use MoveIt with RRTConnect, although other planners may also be used.

To reflect the bidirectional nature of a digital twin, we include a runtime commissioning and monitoring module that closes the loop between the planned and measured trajectories.
The digital twin is continuously synchronized with the physical robot state and uses the same scene representation as planning for online verification.

During execution, the robot controller publishes joint encoder readings at 50--250\,Hz.
We define a synchronization operator
\begin{equation}
  \psi_{\text{sync}}: z(t) \mapsto q^{\text{exec}}(t),
  \label{eq:sync}
\end{equation}
where $z(t)$ denotes the telemetry at time $t$ and $q^{\text{exec}}(t) \triangleq (q_1^{\text{exec}}(t),\ldots,q_R^{\text{exec}}(t))$ is the multi-robot joint configuration.

Given $Q^{\text{plan}}$ and $Q^{\text{exec}}$, we evaluate the deviation metric $\delta(Q^{\text{plan}},Q^{\text{exec}})$ in (\ref{eq:traj_deviation}) and apply the pass/fail rule $V_{\text{traj}}(\cdot,\cdot)$ in (\ref{eq:traj_check}).
Because $Q^{\text{plan}}$ is time-parameterized, we compare $q_r^{\text{exec}}(t)$ and $q_r^{\text{plan}}(t)$ after nearest-neighbor timestamp alignment for each robot $r$.

If $V_{\text{traj}}(Q^{\text{plan}}, Q^{\text{exec}})=\text{fail}$, the system can (i) pause or stop execution, (ii) request human inspection via the digital twin, or (iii) replan the remaining actions from the current robot state.
This makes commissioning auditable and reduces the risk of silently drifting away from the validated plan.

\section{Experiments}
\label{sec:experiments}

\subsection{Experimental Setup}
\label{sec:exp_setup}

\begin{figure}[t]
  \centering
  \includegraphics[width=0.9\columnwidth]{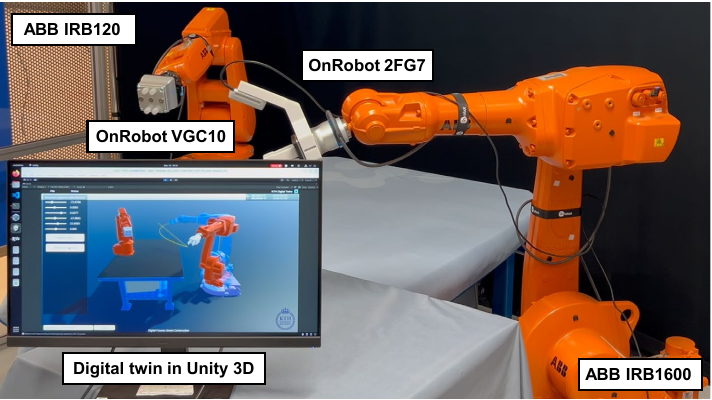}
  \caption{Experimental setup of the dual-arm robotic system.}
  \label{fig:exp_setup}
\end{figure}

The system runs on a dual-arm robotic station comprising an ABB IRB120 (OnRobot VGC10 vacuum gripper) and an ABB IRB1600 (OnRobot 2FG7 parallel gripper), as shown in Fig.~\ref{fig:exp_setup} and detailed in Table~\ref{tab:hardware}. The software stack runs on Ubuntu Linux 22.04 with ROS~2 (Humble), ros2\_control 2.43.1, MoveIt~2, and Unity3D 2021.3.22f1.
All LLM calls use GPT-4o-mini (temperature 0.3).
We use ``end-to-end'' to refer to the complete system pipeline from natural language to physical assembly.

\begin{table}[t]
  \centering
  \caption{Hardware Configuration}
  \label{tab:hardware}
  \setlength{\tabcolsep}{4pt}
  \renewcommand{\arraystretch}{1.1}
  \begin{tabular}{@{}llll@{}}
    \hline
    \textbf{Robot / Gripper} & \textbf{Firmware} & \textbf{Protocol} & \textbf{Update Rate} \\
    \hline
    ABB IRB120         & ABB RobotWare 5.15  & ABB PCSDK      & 50 Hz  \\
    ABB IRB1600        & ABB RobotWare 6.13  & ABB EGM        & 250 Hz \\
    OnRobot 2FG7       & --                  & Modbus         & 2 Hz   \\
    OnRobot VGC10      & --                  & Modbus         & 2 Hz   \\
    \hline
  \end{tabular}
\end{table}

\begin{figure*}[t]
  \centering
  \includegraphics[width=\textwidth]{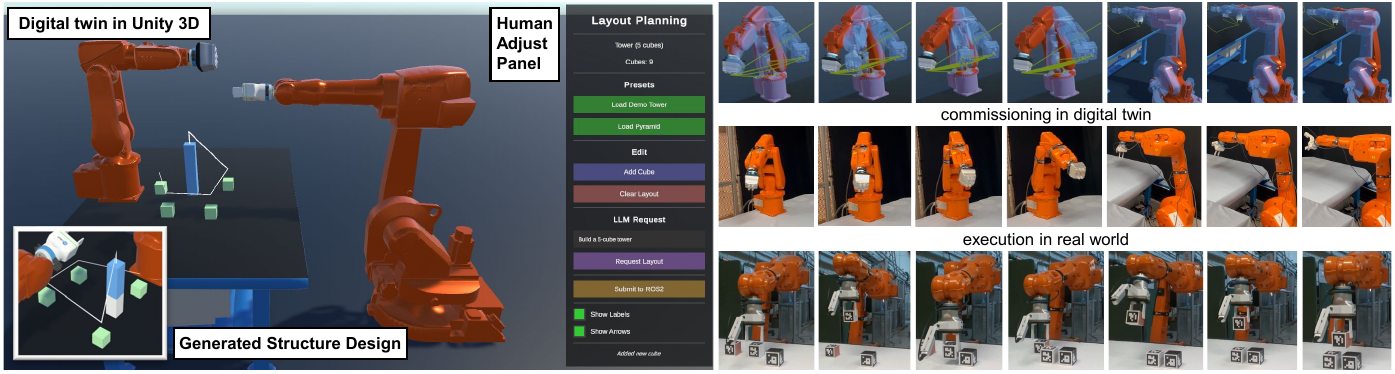}
  \caption{Digital-twin, human-AI interface, robot motion commissioning and execution.}
  \label{fig:dt_interface}
\end{figure*}

We evaluate on 70 natural-language commands organized into five difficulty groups:
A (13 commands, 3--15 steps): well-defined structures with explicit dimensions;
B (15 commands, 5--15 steps): under-specified or vague instructions and patterns requiring semantic grounding;
C (12 commands, 7--16 steps): enclosed and dense structures stressing physical feasibility;
D (18 commands, 6--20 steps): 3D packing and open-ended commands combining scale with ambiguity;
and E (12 commands, 13--25 steps, avg.\ 19.2): scale stress, structural complexity, and recovery stress that exceed Group~D in both step count and constraint density.

\begin{table}[t]
  \centering
  \caption{Summary of Compared Methods}
  \label{tab:method_summary}
  \setlength{\tabcolsep}{2pt}
  \renewcommand{\arraystretch}{1.05}
  \begin{tabular}{@{}llcc@{}}
    \hline
    Method & Paradigm & Skills & Orch. \\
    \hline
    LLM-Direct~\cite{brohan2023can,liang2023code} & Single-pass & -- & -- \\
    LLM-FullPrompt & Prompt eng. & -- & -- \\
    LLM-FixedLoop~\cite{huang2023inner} & Fixed feedback & -- & -- \\
    LLM-AdaptLoop & Adaptive feedback & -- & -- \\
    \hline
    SASS & Tool-use & 1 & -- \\
    SAMS~\cite{attolino2025achieving,capitanelli2024framework} & Tool-use & 8 & -- \\
    \hline
    MultiAgent-RuleOrch~\cite{wu2024autogen,wang2025llm} & Multi-agent & 6 & Rule \\
    Ours-PGEOrch          & Multi-agent & 6   & PGE \\
    Ours-LGOrch           & Multi-agent & 6   & LG \\
    Ours-Hybrid           & Multi-agent & 6+R & PGE+LG \\
    \hline
  \end{tabular}
\end{table}

Table~\ref{tab:method_summary} groups the ten methods in three paradigms. \textit{Non-agentic baselines} (no skill registry) differ in feedback: \textit{LLM-Direct}~\cite{brohan2023can,liang2023code} (one pass), \textit{LLM-FullPrompt} (full $\mathcal{K}$ in prompt), \textit{LLM-FixedLoop}~\cite{huang2023inner} (retry with full feedback), \textit{LLM-AdaptLoop} (retry with violated subset $\phi(E)$, Algorithm~\ref{alg:lazy}). \textit{Single-agent tool-use methods} wrap one LLM in a function-calling loop: Single-Agent Single-Skill (\textit{SASS}, one composite verifier) and Single-Agent Multi-Skill (\textit{SAMS}, eight granular tools)~\cite{attolino2025achieving,capitanelli2024framework}. \textit{Multi-agent methods} share four agents and six skills, differing in command grounding, control flow, and recovery: \textit{MultiAgent-RuleOrch}~\cite{wu2024autogen,wang2025llm} uses raw commands, a hardcoded if-else state machine, and rule-based replan; \textit{Ours-PGEOrch} adds the full Specifier-Designer-Inspector (inspired from PGE) harness with a fixed PGE loop; \textit{Ours-LGOrch} drops the Specifier (raw commands) and instead routes via a LangGraph state machine; \textit{Ours-Hybrid} combines both axes (Specifier $+$ LangGraph) and adds the geometric recovery skills $\xi_{\text{flatten}},\xi_{\text{shift}}$. 
Generation time averages all commands (success and failure alike). Failed runs contribute their full elapsed time, bounded by a 300\,s per-phase timeout and up to 3 replan attempts.

\subsection{Main Results}
\label{sec:main_results}

Table~\ref{tab:main_results} presents success rates across all four difficulty levels and all compared methods. The proposed Ours-Hybrid method is the only one achieving 100\% across all four difficulty levels, outperforming both non-agentic and single-agent baselines.
Among non-agentic methods, LLM-AdaptLoop (93.1\%) outperforms LLM-FixedLoop (82.8\%), confirming that adaptive constraint injection is more effective than a fixed feedback schedule, while SAMS (93.1\%) achieves comparable overall accuracy by benefiting from explicit symbolic verification at each step.
Group~D is the most discriminating: all baselines fall below 90\% (with the weakest reaching only 55.6\%), whereas agentic methods with granular skills sustain 72--89\%, confirming that open-ended, large-scale assembly tasks require the structured planning and multi-agent coordination provided by Ours-Hybrid.

\begin{table}[t]
  \centering
  \caption{Main Results: Success Rate (\%) by Difficulty Level and Generation Time}
  \label{tab:main_results}
  \setlength{\tabcolsep}{2pt}
  \renewcommand{\arraystretch}{1.1}
  \begin{tabular}{@{}lccccccc@{}}
    \hline
    Method & A & B & C & D & Overall & Gen. & Orch. \\
    \hline
    & (13) & (15) & (12) & (18) & (58) & (s) & \\
    LLM-Direct           & 100.0 & 93.3 & 91.7 & 55.6 & 82.8 & 14.9 & -- \\
    LLM-FullPrompt       & 100.0 & 73.3 & 75.0 & 55.6 & 74.1 & 25.1 & -- \\
    LLM-FixedLoop        & 100.0 & 86.7 & 83.3 & 66.7 & 82.8 & 31.5 & -- \\
    LLM-AdaptLoop        & 100.0 & 100.0 & 91.7 & 83.3 & 93.1 & 23.7 & -- \\
    SASS   & 100.0 & 100.0 & 100.0 & 72.2 & 91.4 & 58.6 & -- \\
    SAMS  & 100.0 & 100.0 & 83.3 & 88.9 & 93.1 & 55.9 & -- \\
    Ours-Hybrid          & 100.0 & 100.0 & 100.0 & 100.0 & 100.0 & 85.3 & PGE+LG \\
    \hline
  \end{tabular}
\end{table}

\subsection{Ablation Study}
\label{sec:ablation}

Table~\ref{tab:orchestration_ablation} ablates the multi-agent components across all 70 commands. Ours-PGEOrch is strong on Group~D (94.4\%) where the Specifier disambiguates open-ended commands but drops on Group~C (83.3\%) where over-constrained specifications limit layout flexibility. Ours-LGOrch is the converse, best on C (100\%) but weaker on D (95.5\%) where raw commands lack detail. Ours-Hybrid attains 100\% on all five tiers with the shortest multi-agent generation time (85.3\,s), indicating Specifier grounding upfront reduces replan cycles.

The last row, Ours-Hybrid-LLMCritic, isolates the Inspector axis and directly validates the neuro-symbolic design adopted in this work. Replacing the symbolic Inspector with an equivalently-prompted LLM (52-command subset of Groups~A--D) collapses overall success from 98.1\% to 3.8\%, with 0\% on Groups~C and~D and a 4$\times$ generation-time blowup (85.3\,s$\rightarrow$319.0\,s) caused by unreliable LLM verdicts triggering replan storms. Unlike code review, robotic applications demand exact geometric and kinematic verification with hard numerical thresholds. The generic harness pattern (an LLM commonly fills the Evaluator role) transfers to industrial robotics only when its critic is replaced by a deterministic symbolic verifier. This confirms that the neuro-symbolic split adopted here, which restricts the LLM to language understanding and generation while delegating verification to a symbolic agent, is the architecture that makes the harness effective for this task class.

\begin{table}[t]
  \centering
  \caption{Component Ablation: Success Rate (\%) by Difficulty Level}
  \label{tab:orchestration_ablation}
  \setlength{\tabcolsep}{2pt}
  \renewcommand{\arraystretch}{1.1}
  \begin{tabular}{@{}lcccccccc@{}}
    \hline
    Method & A & B & C & D & E & Overall & Gen. & Msgs \\
    & (13) & (15) & (12) & (18) & (12) & (70) & (s) & \\
    \hline
    MultiAgent-RuleOrch  & 100.0 & 100.0 & 83.3 & 77.8  & 66.7  & 85.7  &  59.4 & 5.6 \\
    Ours-PGEOrch         & 100.0 & 100.0 & 83.3 & 94.4  & 91.7  & 94.3  & 130.3 & 6.5 \\
    Ours-LGOrch          & 100.0 & 100.0 & 100.0 & 95.5 & 91.7  & 97.4  & 125.0 & 6.9 \\
    Ours-Hybrid          & 100.0 & 100.0 & 100.0 & 100.0 & 100.0 & 100.0 & 85.3 & 6.4 \\
    Ours-Hybrid-LLMCritic & 7.1 & 5.9 & 0.0 & 0.0 & -- & 3.8 & 319.0 & 22.0 \\
    \hline
  \end{tabular}
\end{table}

\subsection{Failure Recovery}
\label{sec:failure_recovery}

We evaluate two recovery scenarios: human-initiated modifications in the digital twin (Table~\ref{tab:human_mod}) and deterministic motion-planning failures injected during execution (Table~\ref{tab:determ_failure}).

For human modifications, the operator introduces nine violations spanning four types including workspace boundary (WS), collision (Col), gravity support (Grav), and assembly order (Seq) across structures of 3--21 assembly steps. All methods recover simple cases ($\leq9$ steps). Failures concentrate on complex structures where rule-based replanning collapses (the 12-step dense block, the 21-step canopy workspace violation), with only Ours-LGOrch and Ours-Hybrid succeeding on all nine.

\begin{table}[t]
  \centering
  \caption{Detection and Correction of Invalid Human Modifications}
  \label{tab:human_mod}
  \setlength{\tabcolsep}{2pt}
  \renewcommand{\arraystretch}{1.1}
  \begin{tabular}{@{}lcccccc@{}}
    \hline
    Human Modification & Type & Steps & Rule & PGE & LG & Hyb \\
    \hline
    Move cube to $x{=}1.0$ in tower          & WS   & 3  & \checkmark & \checkmark & \checkmark & \checkmark \\
    Move row to $y{=}0.01$ in wall           & WS   & 9  & \checkmark & \checkmark & \checkmark & \checkmark \\
    Move pillar to $x{=}0.85$ in canopy      & WS   & 21 & $\times$   & $\times$   & \checkmark & \checkmark \\
    Stack two cubes at same position         & Col  & 6  & \checkmark & \checkmark & \checkmark & \checkmark \\
    Overlap cubes in dense block             & Col  & 12 & $\times$   & \checkmark & \checkmark & \checkmark \\
    Remove middle cube from tower            & Grav & 5  & \checkmark & \checkmark & \checkmark & \checkmark \\
    Remove pillar from bridge                & Grav & 5  & \checkmark & \checkmark & \checkmark & \checkmark \\
    Reverse staircase assembly order         & Seq  & 8  & \checkmark & \checkmark & \checkmark & \checkmark \\
    Place roof before pillars in canopy      & Seq  & 21 & \checkmark & \checkmark & \checkmark & \checkmark \\
    \hline
    Total & & & 7/9 & 8/9 & 9/9 & 9/9 \\
    Rate (\%) & & & 77.8 & 88.9 & 100.0 & 100.0 \\
    \hline
  \end{tabular}
\end{table}

For deterministic execution failures, ten blocked regions force structural replanning. Ours-Hybrid combines two mechanisms: dedicated recovery skills ($\xi_{\text{flatten}}, \xi_{\text{shift}}$) bypass the Designer to feed corrected layouts directly to the Inspector, and adaptive Specifier-relaxation reverts to the raw command after repeated failures. MultiAgent-RuleOrch (5/10) handles only workspace and collision failures; Ours-PGEOrch (2/10) is further limited by Specifier over-constraining during recovery. Notably, Ours-LGOrch (10/10) surpasses Ours-Hybrid (9/10) on the narrow-corridor scenario, where the 0.14\,m valid band is too tight for reliable LLM coordinate generation under the expanded specification.

\begin{table}[t]
  \centering
  \caption{Deterministic Failure Recovery (10 Scenarios)}
  \label{tab:determ_failure}
  \setlength{\tabcolsep}{2.5pt}
  \renewcommand{\arraystretch}{1.1}
  \begin{tabular}{@{}lcccccc@{}}
    \hline
    Scenario & Error Type & Rule & PGE & LG & Hyb \\
    \hline
    Workspace boundary ($x{>}0.6$)      & IK failure      & \checkmark & \checkmark & \checkmark & \checkmark \\
    Height limit ($z{>}0.15$)           & Self-collision   & $\times$   & $\times$   & \checkmark & \checkmark \\
    IK singularity (center)             & Singularity      & $\times$   & $\times$   & \checkmark & \checkmark \\
    Env collision ($y{>}0.4$)           & Collision        & \checkmark & $\times$   & \checkmark & \checkmark \\
    Combined (boundary + height)        & Multiple         & \checkmark & \checkmark & \checkmark & \checkmark \\
    Narrow corridor ($x{\in}[0.3,0.4]$) & Constrained      & $\times$   & $\times$   & \checkmark & $\times$ \\
    Low ceiling ($z{>}0.08$)            & Self-collision   & \checkmark & $\times$   & \checkmark & \checkmark \\
    Y-band ($y{\notin}[0.23,0.37]$)     & Env collision    & \checkmark & $\times$   & \checkmark & \checkmark \\
    Diagonal obstacle ($x{+}y{>}0.7$)   & Env collision    & $\times$   & $\times$   & \checkmark & \checkmark \\
    Corner pocket (small area)          & Multiple         & $\times$   & $\times$   & \checkmark & \checkmark \\
    \hline
    Total &  & 5/10 & 2/10 & 10/10 & 9/10 \\
    Rate (\%) &  & 50.0 & 20.0 & 100.0 & 90.0 \\
    \hline
  \end{tabular}
\end{table}

\subsection{Physical Deployment and Real-world Validation}
\label{sec:real_world}

We execute 12 commands through the full motion planning and control pipeline to verify trajectory feasibility and dual-arm task allocation (Table~\ref{tab:moveit_results}).
Fig.~\ref{fig:dt_interface} illustrates the complete commissioning workflow: the Unity3D digital twin and human-AI interface (left) lets the operator issue typed commands and refine the layout by adding, removing, or mouse-dragging cubes; the right panel shows pre-execution preview with a semi-transparent robot model (top), real-robot execution upon human approval (middle), and a completed three-cube pyramid (bottom).

\begin{table}[t]
  \centering
  \caption{Motion Planning Verification (12 Commands, Grouped by Category)}
  \label{tab:moveit_results}
  \setlength{\tabcolsep}{4pt}
  \renewcommand{\arraystretch}{1.1}
  \begin{tabular}{@{}lcccc@{}}
    \hline
    Category (Commands) & Cubes & Arms & Speedup & Pass \\
    \hline
    Simple single-arm (6)   & 3--5 & IRB120        & 1.0$\times$       & 6/6 \\
    Complex single-arm (3)  & 5--6 & IRB120        & 1.0$\times$       & 3/3 \\
    Dual-arm (3)            & 5--6 & IRB120+IRB1600 & 1.0--1.7$\times$ & 3/3 \\
    \hline
    Total                   & 3--6 & --             & up to 1.7$\times$ & 12/12 \\
    \hline
  \end{tabular}
\end{table}

All 12 designs yield valid inverse kinematics and collision-free trajectories, and the three dual-arm commands confirm correct allocation between the IRB120 and IRB1600, achieving up to 1.7$\times$ speedup through parallel execution.
Since trajectory following is a mature problem in industrial robotics, we observed no significant deviation between planned and executed trajectories; the robots reliably tracked the planned paths within nominal control accuracy.
The system outputs an executable Python script in the robot frame for each command, deployable directly on physical hardware without modification.

\section{Discussion}
\label{sec:discussion}

Three findings emerge from the ablation study. First, the Specifier handles ambiguity grounding. Without it, performance drops to 95.5\% on Group~D, confirming that ambiguous commands require structured grounding before generation. Second, LangGraph handles failure-aware workflow. Without it, performance drops to 83.3\% on Group~C, confirming that fixed orchestration cannot adapt to the diverse failure modes of enclosed and dense structures. Third, adaptive routing handles execution-level recovery. Without it, MultiAgent-RuleOrch and Ours-PGEOrch recover only 5/10 and 2/10 of injected failures respectively, while LangGraph routing in Ours-LGOrch and Ours-Hybrid recovers 9--10/10. Dedicated geometric recovery skills ($\xi_{\text{flatten}}, \xi_{\text{shift}}$) provide an additional layer for known geometric failure patterns.

The SDI harness and LangGraph routing address orthogonal aspects of the problem. SDI defines role decomposition (Specifier for language understanding, Designer for structure design generation, Inspector for constraint verification), while LangGraph defines the control flow that routes between agents on failure. Replacing the LLM-based critic of the generic harness pattern with a deterministic symbolic Inspector preserves the harness structure while providing the physical-feasibility guarantees industrial deployment requires. The deterministic failure recovery experiment (Table~\ref{tab:determ_failure}) exposes a Specifier trade-off: the same structure that grounds ambiguous commands over-constrains the search space during recovery. PGEOrch reaches only 2/10 because the Specifier locks in geometric assumptions that conflict with the failure-triggering constraints, and LGOrch (10/10) even surpasses Hybrid (9/10) by exploiting raw commands directly. Hybrid attains 9/10 via an adaptive Specifier-relaxation mechanism that reverts to the raw command after repeated failures. SAMS (93.1\%) does not consistently outperform SASS (91.4\%), suggesting that tool-set size alone is insufficient and that structured multi-agent orchestration is the primary driver of performance on complex tasks. 

As for limitation, the current system assumes known supply-area positions, and visual perception for object localization was not integrated because it is out of the scope of this study. Hybrid averages 85.3\,s per command due to multi-agent communication and LLM API calls, which are time-consuming and expensive; lighter models or cached Specifier outputs may reduce this overhead in latency-sensitive deployments.

\section{Conclusions}
\label{sec:conclusions}

This paper presented an agentic neuro-symbolic planning and commissioning framework for human-in-the-loop industrial robotics.
The proposed system integrates a Specifier-Designer-Inspector (SDI) harness framework, symbolic constraint verification, LangGraph-based dynamic orchestration, domain-specific recovery skills, and a Unity3D digital twin for human review.
Across 70 natural-language commands spanning five difficulty levels, the proposed method achieved 100\% success, showing that symbolic grounding, adaptive routing, and human oversight are all essential for reliable performance on open-ended tasks.
Future work will extend the framework to heterogeneous parts, integrate vision-based localization, and investigate adaptive skill selection to reduce generation time without sacrificing success rate.

\bibliographystyle{IEEEtran}
\bibliography{ref}

@inproceedings{brohan2023can,
  title={Do as {I} can, not as {I} say: Grounding language in robotic affordances},
  author={Brohan, Anthony and Chebotar, Yevgen and Finn, Chelsea and Hausman, Karol and Herzog, Alexander and Ho, Daniel and Ibarz, Julian and Irpan, Alex and Jang, Eric and Julian, Ryan and others},
  booktitle={Conference on Robot Learning},
  pages={287--318},
  year={2023},
  organization={PMLR}
}

@inproceedings{liang2023code,
  title={Code as policies: Language model programs for embodied control},
  author={Liang, Jacky and Huang, Wenlong and Xia, Fei and Xu, Peng and Hausman, Karol and Ichter, Brian and Florence, Pete and Zeng, Andy},
  booktitle={2023 IEEE International Conference on Robotics and Automation (ICRA)},
  pages={9493--9500},
  year={2023},
  organization={IEEE}
}

@inproceedings{huang2023inner,
  title={Inner Monologue: Embodied Reasoning through Planning with Language Models},
  author={Huang, Wenlong and Xia, Fei and Xiao, Ted and Chan, Harris and Liang, Jacky and Florence, Pete and Zeng, Andy and Tompson, Jonathan and Mordatch, Igor and Chebotar, Yevgen and others},
  booktitle={Conference on Robot Learning},
  pages={1769--1782},
  year={2023},
  organization={PMLR}
}

@inproceedings{wu2024autogen,
  title={Autogen: Enabling next-gen {LLM} applications via multi-agent conversations},
  author={Wu, Qingyun and Bansal, Gagan and Zhang, Jieyu and Wu, Yiran and Li, Beibin and Zhu, Erkang and Jiang, Li and Zhang, Xiaoyun and Zhang, Shaokun and Liu, Jiale and others},
  booktitle={First Conference on Language Modeling},
  year={2024}
}

@article{attolino2025achieving,
  title={Achieving Scalable Robot Autonomy via neurosymbolic planning using lightweight local {LLM}},
  author={Attolino, Nicholas and Capitanelli, Alessio and Mastrogiovanni, Fulvio},
  journal={arXiv preprint arXiv:2505.08492},
  year={2025}
}

@article{capitanelli2024framework,
  title={A framework for neurosymbolic robot action planning using large language models},
  author={Capitanelli, Alessio and Mastrogiovanni, Fulvio},
  journal={Frontiers in Neurorobotics},
  volume={18},
  pages={1342786},
  month = jun,
  year={2024},
  publisher={Frontiers Media SA}
}

@article{wang2025llm,
  title={{LLM}-based multi-agent task planning for human-robot collaborative assembly balancing operator experience and efficiency},
  author={Wang, Binbin and Zheng, Lianyu and Wang, Yiwei and Qi, Zhonghua},
  journal={Journal of Manufacturing Systems},
  volume={82},
  pages={1020--1045},
  month = Oct,
  year={2025},
  publisher={Elsevier}
}

@article{garrett2021integrated,
  title={Integrated task and motion planning},
  author={Garrett, Caelan Reed and Chitnis, Rohan and Holladay, Rachel and Kim, Beomjoon and Silver, Tom and Kaelbling, Leslie Pack and Lozano-P{\'e}rez, Tom{\'a}s},
  journal={Annual Review of Control, Robotics, and Autonomous Systems},
  volume={4},
  number={1},
  pages={265--293},
  month = may,
  year={2021},
  publisher={Annual Reviews}
}

@misc{anthropic_harness2025,
  author       = {{Anthropic}},
  title        = {Harness Design for Long-Running Apps},
  howpublished = {Anthropic Engineering Blog},
  year         = {2026},
  url          = {https://www.anthropic.com/engineering/harness-design-long-running-apps}
}

@article{rojek2025RoleGenerative,
  title = {Role of {Generative AI} in {AI}-Based Digital Twins in {Industry} 5.0 and Evolution to {Industry} 6.0},
  author = {Rojek, Izabela and Miko{\l}ajewski, Dariusz and Piszcz, Adrianna and Ma{\l}olepsza, Olga and Kozielski, Miros{\l}aw},
  year = 2025,
  month = jan,
  journal = {Applied Sciences},
  volume = {15},
  number = {18},
  pages = {10102},
  publisher = {Multidisciplinary Digital Publishing Institute},
  issn = {2076-3417},
  doi = {10.3390/app151810102}
}

@inproceedings{lykov2025Industry60,
  title = {Industry 6.0: New Generation of Industry Driven by {{Generative AI}} and Swarm of Heterogeneous Robots},
  shorttitle = {Industry 6.0},
  booktitle = {2025 {{IEEE}}/{{RSJ International Conference}} on {{Intelligent Robots}} and {{Systems}} ({{IROS}})},
  author = {Lykov, Artem and Cabrera, Miguel Altamirano and Konenkov, Mikhail and Serpiva, Valerii and Gbagbe, Koffivi Fid{\`e}le and Alabbas, Ali and Fedoseev, Aleksey and Moreno, Luis and Khan, Muhammad Haris and Guo, Ziang and Tsetserukou, Dzmitry},
  year = 2025,
  month = oct,
  pages = {19992--19997},
  issn = {2153-0866},
  doi = {10.1109/IROS60139.2025.11247291}
}

@article{liu2026ConstrucTwinDigital,
  title = {{{ConstrucTwin}}: Digital Twin-Driven Multirobot Construction System Toward {Industry 5.0}},
  shorttitle = {{{ConstrucTwin}}},
  author = {Liu, Zhihao and Silva, Jorge and Zhong, Ruirui and Qin, Qiang and Roy, Neelabhro and {Nan Fernandez-Ayala}, Victor and Lesko, Johan and H{\aa}kansson, Ulf and Sandberg, Sara and Dimarogonas, Dimos V. and Gross, James and Vincent Wang, Xi and Wang, Lihui},
  year = 2026,
  month = apr,
  journal = {IEEE Transactions on Systems, Man, and Cybernetics: Systems},
  volume = {56},
  number = {4},
  pages = {2924--2939},
  issn = {2168-2216, 2168-2232},
  doi = {10.1109/TSMC.2026.3658622}
}

@article{ni2025LargeLanguage,
  title = {A Large Language Model-Based Manufacturing Process Planning Approach under {Industry 5.0}},
  author = {Ni, Mingzhe and Wang, Tao and Leng, Jiewu and Chen, Chong and Cheng, Lianglun},
  year = 2025,
  month = feb,
  journal = {International Journal of Production Research},
  pages = {1--20},
  publisher = {Taylor \& Francis},
  issn = {0020-7543},
  doi = {10.1080/00207543.2025.2469285}
}

@article{dong2025NextgenerationLLM,
  title = {Towards a Next-Generation {{LLM}} Empowered Low-Code Programming Industrial Robotic System for Human-Centric Smart Manufacturing},
  author = {Dong, Wenhang and Li, Dongpeng and Ji, Yuchen and Chen, Hongpeng and Liu, Shimin and Ma, Zheng and Hao, Fang and Ji, Yuqi and Xing, Hongwen and Zheng, Pai},
  year = 2025,
  month = dec,
  journal = {Journal of Manufacturing Systems},
  volume = {83},
  pages = {675--686},
  issn = {0278-6125},
  doi = {10.1016/j.jmsy.2025.10.012}
}

@article{fan2025EmbodiedIntelligence,
  title = {Embodied Intelligence in Manufacturing: Leveraging Large Language Models for Autonomous Industrial Robotics},
  shorttitle = {Embodied Intelligence in Manufacturing},
  author = {Fan, Haolin and Liu, Xuan and Fuh, Jerry Ying Hsi and Lu, Wen Feng and Li, Bingbing},
  year = 2025,
  month = feb,
  journal = {Journal of Intelligent Manufacturing},
  volume = {36},
  number = {2},
  pages = {1141--1157},
  issn = {1572-8145},
  doi = {10.1007/s10845-023-02294-y}
}

@article{kadri2025LLMdrivenAgent,
  title = {{{LLM-driven}} Agent for Speech-Enabled Control of Industrial Robots: {{A}} Case Study in Snow-Crab Quality Inspection},
  shorttitle = {{{LLM-driven}} Agent for Speech-Enabled Control of Industrial Robots},
  author = {Kadri, Ibrahim and Selouani, Sid Ahmed and Ghribi, Mohsen and Ghali, Rayen and Mekhoukh, Sabrina},
  year = 2025,
  month = sep,
  journal = {Results in Engineering},
  volume = {27},
  pages = {106660},
  issn = {2590-1230},
  doi = {10.1016/j.rineng.2025.106660}
}

@article{li2026EnvironmentDrivenLLMGuided,
  title = {Environment-Driven and {LLM}-Guided Multi-Robot Task Inference and Allocation Under Temporal Logic Specifications},
  author = {Li, Lin and Chen, Ziyang and Kan, Zhen},
  month = jan,
  year = 2026,
  journal = {IEEE Transactions on Automation Science and Engineering},
  volume = {23},
  pages = {2925--2940},
  issn = {1545-5955, 1558-3783},
  doi = {10.1109/TASE.2026.3659055}
}

@article{tsushima2025TaskPlanninga,
  title = {Task Planning for a Factory Robot Using Large Language Model},
  author = {Tsushima, Yosuke and Yamamoto, Shu and Ravankar, Ankit A and Luces, Jose Victorio Salazar and Hirata, Yasuhisa},
  year = 2025,
  month = mar,
  journal = {IEEE Robotics and Automation Letters},
  volume = {10},
  number = {3},
  pages = {2383--2390},
  issn = {2377-3766},
  doi = {10.1109/LRA.2025.3531153}
}

@article{wang2026LLMMTMPLarge,
  title = {{{LLM-MTMP}}: {{A}} Large Language Model-Based Multi-Agent Task and Motion Planning Framework for Power Inspection Robots},
  shorttitle = {{{LLM-MTMP}}},
  author = {Wang, Zongyuan and Zhou, Xin and Mao, Jianliang and Zhang, Chuanlin and Cui, Chenggang and Yang, Jun},
  year = 2026,
  month = jan,
  journal = {Journal of Industrial Information Integration},
  volume = {49},
  pages = {101014},
  issn = {2452-414X},
  doi = {10.1016/j.jii.2025.101014}
}

@article{xia2025LeveragingLarge,
  title = {Leveraging Large Language Models to Empower {Bayesian} Networks for Reliable Human-Robot Collaborative Disassembly Sequence Planning in Remanufacturing},
  author = {Xia, Liqiao and Hu, Youxi and Pang, Jiazhen and Zhang, Xiangying and Liu, Chao},
  year = 2025,
  month = apr,
  journal = {IEEE Transactions on Industrial Informatics},
  volume = {21},
  number = {4},
  pages = {3117--3126},
  issn = {1941-0050},
  doi = {10.1109/TII.2024.3523551}
}

@inproceedings{rashed2025ai,
  title={AI-Driven Multi-Agent Demand Response Framework for Residential Load Optimization Using {CrewAI}},
  author={Rashed, Ghamgeen Izat and Bahageel, Ahmed OM and Gony, Hashim Ali I and Badjan, Ansumana and Shaheen, Husam I},
  booktitle={2025 IEEE 20th Conference on Industrial Electronics and Applications (ICIEA)},
  pages={1--10},
  year={2025},
  organization={IEEE}
}

@inproceedings{mandulapalli2025development,
  title={Development of Agentic Workflows with {LangGraph} for Software Development Life Cycle Automation},
  author={Mandulapalli, Shriraj and Hernandez, Emilio and Hall, Wayne Jordan and Chakeri, Alireza and Jaimes, Luis},
  booktitle={North American Conference on Industrial Engineering and Operations Management-Computer Science Tracks},
  pages={45--54},
  year={2025},
  organization={Springer}
}

@article{hitzler2022NeurosymbolicApproaches,
  title = {Neuro-Symbolic Approaches in Artificial Intelligence},
  author = {Hitzler, Pascal and Eberhart, Aaron and Ebrahimi, Monireh and Sarker, Md Kamruzzaman and Zhou, Lu},
  year = 2022,
  month = jun,
  journal = {National Science Review},
  volume = {9},
  number = {6},
  pages = {nwac035},
  issn = {2095-5138},
  doi = {10.1093/nsr/nwac035}
}

@article{singh2026NeuralsymbolicGrammatical,
  title = {Towards Neural-Symbolic Grammatical Inference for Endangered Languages Using Integrating Graph Neural Networks and Instruction-Tuned Language Models},
  author = {Singh, Manu and Gupta, Neha and Tyagi, Shiva and Rani, Ashima and Kumar, Vinod and Sharma, Surbhi},
  year = 2026,
  month = mar,
  journal = {Engineering Applications of Artificial Intelligence},
  volume = {168},
  pages = {114011},
  issn = {0952-1976},
  doi = {10.1016/j.engappai.2026.114011}
}

@article{mao2026BuildingIntelligent,
  title = {Building Intelligent Agents with Neuro-Symbolic Concepts},
  author = {Mao, Jiayuan and Tenenbaum, Joshua B. and Wu, Jiajun},
  year = 2026,
  month = jan,
  journal = {Commun. ACM},
  volume = {69},
  number = {2},
  pages = {69--79},
  issn = {0001-0782},
  doi = {10.1145/3715316}
}

@article{wang2026FunctionCalling,
  title = {Function Calling in Large Language Models: Industrial Practices, Challenges, and Future Directions},
  shorttitle = {Function {{Calling}} in {{Large Language Models}}},
  author = {Wang, Maolin and Zhang, Yingyi and Yu, Bowen and Hao, Bingguang and Peng, Cunyin and Chen, Yicheng and Zhou, Wei and Gu, Jinjie and Zhuang, Chenyi and Guo, Ruocheng and Wang, Wanyu and Zhao, Xiangyu},
  year = 2026,
  month = feb,
  journal = {ACM Comput. Surv.},
  volume = {58},
  number = {9},
  pages = {238:1--238:37},
  issn = {0360-0300},
  doi = {10.1145/3788284}
}

@article{liu2025establishment,
  title={Establishment and Synchronisation of Digital Twins for Multi-robot Systems in Manufacturing},
  author={Liu, Zhihao and Liu, Sichao and Wang, Tianyu and Wang, Lihui and Wang, Xi Vincent},
  journal={Procedia CIRP},
  volume={134},
  pages={419--424},
  year={2025},
  publisher={Elsevier}
}

\end{document}